\documentclass[conference,compsoc]{IEEEtran}
\ifCLASSOPTIONcompsoc
  \usepackage[nocompress]{cite}
\else
  \usepackage{cite}
\fi
\ifCLASSINFOpdf
\else
\fi
\usepackage{cite}
\usepackage{url}
\usepackage{amsmath,amssymb,amsfonts}
\usepackage{algorithmic}
\usepackage{graphicx}
\usepackage{epsfig}
\usepackage{textcomp}
\usepackage{xcolor}
\usepackage{textpos}
\usepackage[normalem]{ulem}

\begin{document}

\title{Simulating Synchrony Loop Networks in the Open Source RISP Neuroprocessor}
\author{
\IEEEauthorblockN{Jackson Mowry}
\IEEEauthorblockA{
EECS Department\\
University of Tennessee\\
Knoxville, TN USA\\
jmowry4@vols.utk.edu, Orcid 0009-0005-8290-3194\\
}

\and
\IEEEauthorblockN{Patrick Abbs}
\IEEEauthorblockA{
Cambrya, Inc.\\
Austin, TX USA\\
patrick@cambrya.co, Orcid 0009-0006-9353-5056\\
}
}

\title{Simulating Synchrony Loop Networks in the Open Source RISP Neuroprocessor}

\maketitle

\begin{abstract}
Neuromorphic spiking neural networks (SNNs) offer a promising alternative to conventional deep neural networks for tasks with computational resource or data constraints. However, their practical applications have been limited by comparatively weak performance on complex learning tasks. Experimental approaches such as Synchrony Loop Propagation (SLP) increasingly seek to address this problem through more sophisticated and heterogeneous neuron models, and have achieved encouraging initial results. However, these neuron models do not readily translate to standard neuromorphic systems designed to support simple leaky integrate-and-fire neurons.

We present an implementation of an SLP network on the RISP neuroprocessor, an event-driven neuromorphic simulation platform, and evaluate its performance on an unsupervised musical instrument clustering task. The network achieves clustering accuracy comparable to the DBSCAN algorithm, while providing over an order of magnitude improvement in runtime speed and computational efficiency relative to a prior non-neuromorphic SLP implementation.

These results demonstrate that SLP’s core mechanisms can be effectively translated into a neuromorphic architecture to support complex unsupervised learning. More broadly, this work highlights the potential of heterogeneous and extensible neuron models to expand the design space of neuromorphic systems to more complex learning tasks.

\end{abstract}


\section{Introduction}
Deep neural networks (DNNs) have achieved remarkable success, but their high computational costs and extensive data requirements present problematic limitations. Neuromorphic spiking neural networks (SNNs) offer an alternative through distributed, event-driven computation inspired by biological systems. In SNNs, neurons act as autonomous units that perform localized computations information only when and as needed, yielding substantial gains in efficiency and latency. However, SNNs still lag behind DNNs in learning performance on complex tasks, restricting their practical applicability.

One possible contributing factor to SNN learning limitations is a relative lack of neuron model heterogeneity. Biological neural systems encompass wide ranges of neuron- and synapse-specific mechanisms, including diverse neurotransmitter types, dynamic forms of short-term plasticity, and bidirectional signaling pathways. In contrast, most SNN frameworks support homogeneous networks that primarily rely on forward spikes and long-term weight updates~\cite{s23063037}. Incorporating richer, more heterogeneous neuron and synapse models may provide a pathway to more effective learning capabilities.

A second contributing factor may be an overemphasis on neurons as independent learning units. Biological systems exhibit sophisticated population-level dynamics that arise from decentralized local interactions, suggesting that neuron and synapse mechanisms may be designed to shape emergent coordinated network activity in addition to governing individual neuron behaviors. Although global dynamics do emerge from existing SNN learning frameworks~\cite{10.1371/journal.pcbi.1011006}, these effects are typically indirect byproducts of local learning rules rather than intentionally engineered population coordination mechanisms. Designing localized mechanisms that directly promote targeted larger-scale dynamics represents a promising and underexplored direction for SNNs.

In this work, we implement an experimental learning framework, Synchrony Loop Propagation (SLP), on the RISP neuroprocessor. SLP introduces continuous unsupervised bidirectional synaptic feedback loops, which produce coordinated population-level learning dynamics under strictly localized signaling constraints. SLP also adopts a flexible heterogeneous network architecture in which neuron types are specialized for distinct functional roles. Prior work shows that SLP can achieve strong performance on tasks such as blind sound source separation (BSS)~\cite{10766538}.

However, conventional neuromorphic systems are not designed to fully support the mechanisms required for SLP. As a result, early SLP networks were implemented on non-neuromorphic software that incurred substantial computational overhead and runtime latency, rendering SLP unsuitable for deployment to real-time edge processing environments such as hearing aids.

The RISP neuroprocessor, part of the TENNLab open-source framework~\cite{pzg:22:risp,pdg:24:risp}, provides a flexible event-driven neuromorphic simulation environment. Although originally designed for homogeneous integrate-and-fire neurons, RISP can be extended to support more complex neuron and synapse models, and offers substantially improved computational efficiency relative to prior SLP implementations.

Motivated by these considerations, in this work we extend RISP to support key SLP mechanisms, and translate a subset of layers from an existing SLP BSS network into RISP to function as a standalone sound profile clustering network. We evaluate the resulting system on an unsupervised musical instrument clustering task, comparing its accuracy to the DBSCAN clustering algorithm, and its processing speed to the original SLP implementation.

Results show that RISP supports all translated SLP mechanisms and achieves clustering performance comparable to DBSCAN, while providing over a 50$\times$ runtime speed improvement that surpasses real-time processing. These findings demonstrate that RISP is a viable and efficient platform for implementing complex, heterogeneous neuromorphic systems, and lays the groundwork for more sophisticated tasks and network designs in future experiments.

\section{Related Work}
Previous works have explored using neuromorphic systems, specifically SNNs, to perform unsupervised clustering. One such system is presented in~\cite{diamond2019unsupervised}, in which STDP-based learning and lateral inhibition allow for the clustering of arbitrary data. Lateral inhibition allows for forming cluster centers with larger separation through the local competition among a population of neurons. STDP learning was used to set weights, however the training method required knowing ahead of time how many cluster centers should be fit. In addition, their learning rule sent spikes to the ``correct'' output neuron for each cluster, requiring information that is not available in a truly unsupervised environment.  

\section{Synchrony Loop Propagation}

This section introduces the core mechanisms of the Synchrony Loop Propagation (SLP) learning framework and develops an intuition for how localized interactions give rise to emergent population-level learning dynamics. Implementation details, including exact equations, are provided in Section \ref{sec:risp_slp_implementation}.

Like spike-timing-dependent plasticity (STDP), SLP is at its core an unsupervised learning framework that leverages temporal spikes to drive synaptic plasticity. However, in contrast to STDP, synaptic information in SLP modifies presynaptic output routing in addition to postsynaptic input integration. This coupling creates a continuous, bidirectional feedback process in which local interactions propagate rapidly across the network, giving rise to coordinated global behaviors that learn from unstructured sensory inputs.

\subsection{Postsynaptic Prediction} \label{sec:postsynaptic_prediction}
Because learning in SLP emerges from population-level dynamics, it cannot be characterized at the level of individual neurons in isolation. Instead, we will examine the complementary roles of postsynaptic and presynaptic processes to see how they interact, beginning with postsynaptic prediction. In conventional neuron models, synaptic weights act as scaling multipliers on incoming signals. In SLP, input synapses instead encode temporal predictions of input signals without directly modifying them. 

Each synapse maintains two state variables: a long-term \textit{\textbf{strength}} property and a short-term \textit{\textbf{pressure}} variable. The strength property encodes a stable representation of stimulus structure. SLP mechanisms are heterogeneous, and neuron types can employ different strength encodings for different stimulus types (e.g. images, sounds, event sequences). In the formulation considered here, synaptic strengths define a normalized distribution of expected input spike rates across input synapses. For example, a 3-synapse neuron with strengths $\{10.0, 50.0, 40.0\}$ corresponds to an expected relative spike distribution of $\{0.1, 0.5, 0.4\}$. 

The pressure variable provides a dynamic, real-time prediction of incoming spike timing. At any given moment, the relative pressures across a neuron’s input synapses define a probability-like distribution of the expected synaptic source of the next spike. For example, pressures $\{0.5, 0.9, 1.6\}$ express a prediction that the next spike is most likely to arrive through synapse 3, and least likely to arrive through synapse 1.

Upon receiving an input spike, a neuron updates its prediction by redistributing pressure across all input synapses. Specifically, pressure equal to the magnitude of the received spike is removed from the receiving synapse and redistributed across all synapses in proportion to their strengths (Fig \ref{fig:slp_core_mechanisms}b). The result of this process is that each synapse has a pressure inflow corresponding to its predicted spike rate relative to other synapses, and outflow corresponding to its actual relative spike rate. 

For example, the neuron with strengths $\{10.0, 50.0, 40.0\}$ will allocate 10\% of the pressure displaced by each received spike to synapse 1. This means that pressure inflow and outflow for synapse 1 will equalize when the synapse receives every 10th input spike for the neuron. If the synapse’s relative spike rate is higher than 10\%, it will not replenish enough pressure after a received spike to anticipate the next spike before it arrives. If it is lower than 10\%, it will predict its next spike too early at the cost of predictive accuracy for the other synapses. 

Thus, pressure dynamics encode prediction accuracy. When a neuron’s strength distribution matches the true input spike rate distribution, pressure concentrates on the correct synapse immediately prior to each spike. When the representation is inaccurate, this predictive alignment degrades. Postsynaptic pressure thereby serves as a local measure of how well a neuron models its input.

\subsection{Presynaptic Routing} \label{sec:presynaptic_routing}
Postsynaptic prediction forms one half of a feedback loop that is completed by presynaptic routing, whereby presynaptic neurons use postsynaptic pressure predictions to preferentially route spikes toward neurons that accurately predict spike timings.
As in LIF neuron models, a presynaptic SLP neuron emits a spike when its accumulated input charge exceeds a threshold. However, instead of the neuron sending a uniform-magnitude spike across all output synapses, it instead sends an individualized graded spike through each synapse whose magnitude depends on the synapse’s pressure value at the time of firing.

\begin{figure}
    \centering
    \includegraphics[width=1\linewidth]{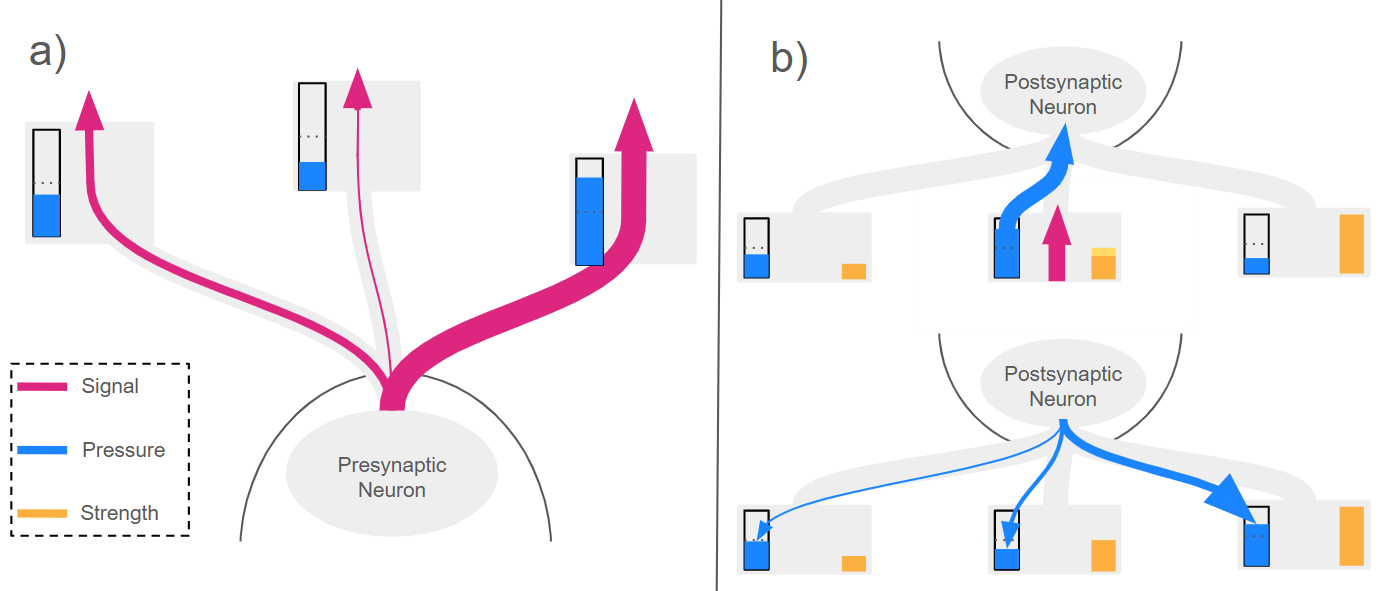}
    \caption{a) Upon firing an output spike, a presynaptic neuron distributes the spike across synapses proportionally to pressure. b) Upon receiving an input spike, a postsynaptic neuron increases the strength of the receiving synapse and redistributes pressure from the synapse across all synapses proportionally to strength.}
    \label{fig:slp_core_mechanisms}
\end{figure}

The computations governing this graded spike routing are heterogeneous across neuron types, but in its basic formulation the total spike magnitude is normalized to 1.0 and distributed across output synapses in proportion to their pressures (Fig \ref{fig:slp_core_mechanisms}a). This weighted allocation biases spike transmission toward synapses with higher pressure, i.e., toward postsynaptic neurons that more strongly predict the spike.

This routing mechanism reinforces predictive accuracy. Strongly predictive neurons consistently anticipate incoming spikes with high synaptic pressure concentrations, thereby attracting larger portions of each spike that in turn further amplify pressure redistribution. Conversely, poorly predictive neurons attract smaller spike portions which then further depress pressure redistribution.

Together, postsynaptic prediction and presynaptic routing form a coordinated yet decentralized population-level feedback loop that progressively concentrates stimulus-driven activity into neurons that best represent the underlying input structure, and away from neurons that represent other stimuli. 

\subsection{Emergent Learning} \label{sec:emergent_learning}
The final component of the SLP feedback loop is the learning mechanism that shapes postsynaptic synaptic strength distributions to represent input stimuli. This mechanism is computationally simple: whenever a postsynaptic neuron receives an input spike, the strength of the corresponding synapse is incremented proportionally to the magnitude of the received signal (Fig \ref{fig:slp_core_mechanisms}b).

The effect of this rule is best understood in the context of SLP’s emergent routing dynamics. At initialization, neurons possess randomly distributed low-magnitude synaptic strengths. When a stimulus is presented, its input spike distribution will align more closely with some neurons than others, even if this initial alignment is weak. Guided by pressure-driven feedback, presynaptic routing then concentrates high-magnitude spikes toward the neuron with the highest relative predictive accuracy, and low-magnitude spikes to all other neurons. As a result, synaptic strength updates are localized primarily to this neuron, causing its strength distribution to adapt toward the observed input pattern. Repeated exposure reinforces this alignment, and the neuron progressively specializes to represent the stimulus.

When a different stimulus is introduced, the synaptic pressure predictions of the previously learned neuron rapidly diverge from the actual timing of the input spikes, and presynaptic neurons correspondingly reroute spikes to a more predictive neuron. The original neuron becomes inactive until the original stimulus reappears, at which point its learned representation again attracts activity. In this way, SLP produces a form of competitive, unsupervised specialization in which neurons self-organize to represent distinct input patterns.

When performing instrument clustering an SLP network encodes a structural representation of the temporal distribution of spikes unique to each instrument. This translation from a rate coding of an instrument's harmonic distribution into synaptic strengths enables online learning, during which many stimuli can be observed, with learning occurring gradually over time. 

\subsection{SLP Extension and Modulation}
The mechanisms described above form a core foundation for emergent learning in SLP that can be extended and modified to support task-specific functionality. Analogous to the diversity of cell types in biological systems, SLP neuron models may incorporate a wide variety of heterogeneous specialized behaviors, provided they adhere to neuromorphic locality constraints. This subsection outlines representative extensions used in an SLP network designed for instrument clustering.

\subsubsection{Continuous Signal Release} \label{sec:continuous_release}
For most SLP neuron types, output signals are emitted as spikes that are fired when accumulated input exceeds a threshold. However, for neurons that must represent fine-grained temporal structure, such as sub-millisecond auditory features, discrete spikes can be problematically slow and inexact.

In these cases, neurons may instead emit a continuous output signal proportional to their input excitation at each time step. This improves temporal precision and can accelerate the underlying pressure prediction / signal routing feedback loop for rapidly evolving stimuli.

This design is inspired by cochlear hair cells in biological auditory systems, which encode frequency amplitude and phase information via continuous neurotransmitter release rather than more typical action potential spikes~\cite{GOUTMAN20153354}. Similarly, the input layer of the SLP auditory network decomposes incoming signals into frequency components and communicates this information to downstream layers using continuous real-time signaling.

\subsubsection{Representational Maturity} \label{sec:maturity}
It is often desirable for neuron synaptic strengths to transition from early plasticity to later stability, to ensure that learned representations are retained without risk of being overwritten. In biological neurons, one mechanism supporting this transition is LTP saturation~\cite{ltpsaturation}, where incremental increases in synaptic strength diminish as aggregate synaptic strength across input synapses approaches an upper bound.

Inspired by this behavior, neurons designed to learn sounds in the SLP network incorporate a \textit{\textbf{maturity}} variable defined as the ratio of total input synaptic strength to a fixed upper limit. This maturity term is then applied as an inverse scaling multiplier on synaptic strength increments, such that strength changes gradually slow as a neuron develops. As a result, neurons undergo a transition from initial plasticity to stable encoding, retaining learned stimulus representations while remaining capable of limited refinement.

\subsubsection{Multiple Signal Types} \label{sec:signal_types}
SLP neurons can also support multiple signal types that influence neuron and synaptic behavior in distinct ways, analogous to the diversity of neurotransmitters in biological systems. In the auditory SLP network, excitatory signaling in some layers is decomposed into two components: an \textit{\textbf{exploratory}} signal and an \textit{\textbf{activating}} signal. Only activating signals contribute to long-term synaptic strength modification, while exploratory signals modulate whether such modification is permitted.

This mechanism is inspired by the interaction between AMPA and NMDA receptors in biological neurons. In that system, sodium influx through AMPA receptors does not directly trigger long-term potentiation (LTP), but facilitates NMDA receptor calcium permeability if sufficient sodium-induced depolarization is achieved~\cite{doi:10.1126/science.1382314}. Calcium influx then triggers LTP.

Analogously, some types of SLP postsynaptic neurons accept activating signals only after accumulating a threshold level of exploratory input. This gating mechanism restricts synaptic strength updates to neurons that consolidate concentrated signal influx for a stimulus, fully protecting the synaptic strength profiles of all other neurons that do not match a stimulus.

\subsubsection{Neuron Subpopulation Coordination} \label{sec:coordinator}
Some SLP network layers benefit from lateral coordination of neuron activity, which can be governed by a non-spiking “coordinator” node inspired by biological astrocytes. Astrocytes are non-spiking glial cells that often help regulate the synaptic plasticity of neuron sub-populations by controlling the availability of co-agonists required for receptor activation. For example, NMDA receptors activation depends not only on postsynaptic depolarization but also on the presence of glycine or D-serine, which are often regulated by astrocytes~\cite{PANATIER2006775}.

The auditory SLP network adopts a similar strategy to govern competition within its final clustering layer. A coordinator node connected to each neuron in the layer monitors exploratory signal accumulations across neurons, and selectively enables activating signal absorption for the most strongly responding neuron at each time step. This augments the neuron self-gating process described earlier, effectively imposing an additional winner-take-all mechanism whereby a stimulus can only activate a single neuron to ensure mutual exclusivity in stimulus classification.

\section{RISP SLP Implementation} \label{sec:risp_slp_implementation}
Several mechanisms introduced by SLP are not supported by standard neuromorphic architectures, so early SLP networks were implemented outside of, and without the efficiency benefits of, neuromorphic systems. The RISP neuroprocessor is an open-source framework that includes a simulator for running neuromorphic networks in software. The simulator was originally designed to support homogeneous LIF-based SNNs, but its neuron and synapse models are flexible beyond requiring  that neuron behavior is localized and event-based. 

Thus, RISP provides a suitable framework for testing the viability of implementing an SLP network in a neuromorphic environment. The RISP SLP implementation described in this section sought to answer two questions. First, are SLP mechanisms supportable under neuromorphic constraints, and second, does translating an SLP network into a neuromorphic simulator yield significant improvements in processing efficiency and memory utilization?

To answer these questions, we translated a heterogeneous five-layer SLP network into RISP. This network originally functioned as a subset of layers within a larger SLP network designed to perform blind sound source separation (BSS), but was adapted into a standalone network that performs clustering of harmonic sound sources such as musical instruments. RISP’s event-driven architecture and spike queue data structure were retained, with support added for heterogeneous networks as well as for five distinct neuron types, each described below. 

\subsection{Frequency Neurons} \label{sec:frequency_neurons}
The network’s input layer consists of frequency neurons modeled after cochlear hair cells, which simulate continuous release by emitting a graded output spike each time step. Each neuron represents a frequency from 10Hz to 24kHz separated by 10Hz increments, and each time step receives a graded input spike with a value corresponding to its current frequency amplitude. This spike is immediately converted into an output spike with an equivalent total magnitude, which is distributed across synapses proportionally to synaptic pressure (Eq. \ref{eq:frequency_routing}) as described in Section \ref{sec:presynaptic_routing}.

\begin{equation}
    SI_{i} = \frac{PR_{i}}{\sum{}_{j}PR_{j}} 
    \label{eq:frequency_routing}
\end{equation}

This routing interacts with postsynaptic pressure dynamics in the subsequent pitch layer to group frequencies belonging to the same sound source.

\subsection{Pitch Neurons} \label{sec:pitch_neurons}
Pitch neurons in the network’s second layer act as feature extractors. Rather than learning new patterns, they are preconfigured to identify instances of a common type of sound called a harmonic series, and to extract characteristics for later layers to learn from. A harmonic series is composed of frequencies that are all multiples of a shared fundamental frequency \textit{\textbf{ff}}. For example, a 100 ff series contains frequencies $\{100, 200, 300, …\}$Hz. Most sounds that have a note or pitch are harmonic series, including voices and musical instruments.

Each pitch neuron is tuned to a particular fundamental by initializing connections only from frequency neurons that make up its series. For example, the 100 ff neuron has connections from the $\{100, 200, 300, …\}$Hz neurons, as well as lower-strength connections from proximal neurons such as $\{95, 105, 195, 205, …\}$Hz. 

\begin{figure}
    \centering
    \includegraphics[width=1\linewidth]{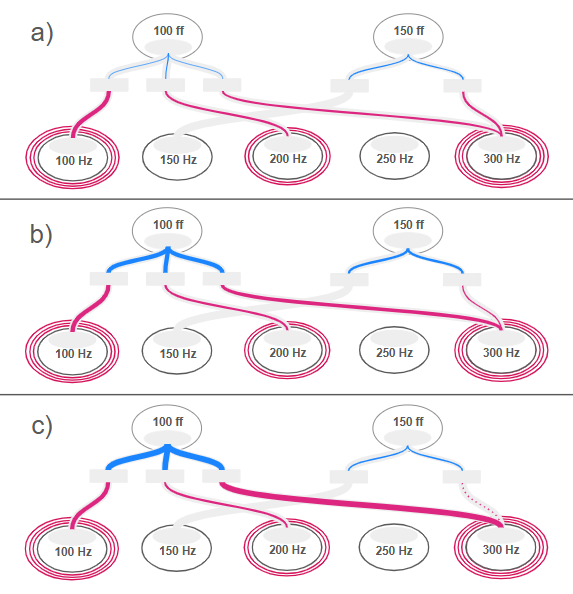}
    \caption{a) A harmonic stimulus appears while all pitch neurons are at rest with pressures set to the minimum baseline. This causes the 300Hz neuron to split output signal routing equally across both pitch neurons. b) Signal inflow causes isq (and therefore exerted pressure) to increase for both pitch neurons, but at a greater rate for the 100 ff neuron. c) This initiates a positive feedback loop that consolidates signal routing to the 100 ff neuron for as long as the stimulus persists.}
    \label{fig:pitch_pressure}
\end{figure}

The pitch layer is designed to consolidate routing of all frequency signals for a harmonic series stimulus to a single corresponding pitch neuron (Fig \ref{fig:pitch_pressure}), which is accomplished through a specialized form of postsynaptic pressure allocation. Instead of redistributing a fixed pool of pressure as described in Section {\ref{sec:postsynaptic_prediction}}, pitch neurons track a recency-weighted average input signal quantity \textit{\textbf{isq}} (Eq. \ref{eq:isq}) and set the pressure of each synapse to a small baseline value \textit{b} plus isq scaled by the synapse’s fixed-value strength (Eq. \ref{eq:pitch_pressure}). Strength is higher for synapses from frequencies \textit{f} that correspond to a lower harmonic \textit{h} and that are closer to their harmonic’s frequency \textit{hf} (Eq. \ref{eq:pitch_strength})

\begin{equation}
    isq_{t+1} = (decay * isq_{t}) + ((1 - decay) * SI_{t})
    \label{eq:isq}
\end{equation}

\begin{equation}
    PR_{i} \leftarrow{} isq * ST_{i} + b
    \label{eq:pitch_pressure}
\end{equation}

\begin{equation}
    ST_{i} = 0.8^{h_{i}} * \left(1 - \frac{|f_{i} - hf_{i}|}{10}\right)
    \label{eq:pitch_strength}
\end{equation}

This creates a positive feedback loop between signal inflow rate and exerted pressure, such that the pitch neuron with the highest-activity harmonic series will consolidate near-complete signal routing from every frequency in its series for as long as the series is present.

Pitch neurons also perform specialized output signal routing, which unlike most SLP neuron types does not incorporate synaptic pressure. Instead, input signal amounts are compartmentalized by harmonic and forwarded to corresponding downstream neurons. For example, a 100 ff neuron pools signals from $\{95, 100, 105\}$Hz synapses into a harmonic 1 compartment, and $\{195, 200, 205\}$Hz synapse signals into a harmonic 2 compartment. When total input signal exceeds a firing threshold, each signal pool is routed to a dedicated output synapse that transmits to a corresponding harmonic neuron in the next layer (Eq. \ref{eq:pitch_output}).

\begin{equation}
    SI_{i} = SIh_{i}
    \label{eq:pitch_output}
\end{equation}

\subsection{Harmonic Neurons} \label{sec:harmonic_neurons}
Harmonic neurons each map to a particular harmonic between 1 and 5, and collectively construct a feature profile representing the relative amplitudes of each harmonic in a sound. Harmonic profiles are analogous to fingerprints, varying significantly between different sounds but remaining largely consistent for any particular sound even when it changes pitch. They are what make a clarinet sound like a clarinet and a violin sound like a violin, no matter what note they play.

Each neuron passively integrates the input signals forwarded to it without using pressure or strength mechanisms (Eq. \ref{eq:harmonic_signal}), because pitch neurons’ self-contained output signal routing maps and transmits a stable and normalized harmonic profile for a stimulus even if the stimulus pitch changes. 

\begin{equation}
    SI  \leftarrow{} SI + SI_{i}
    \label{eq:harmonic_signal}
\end{equation}

The harmonic layer firing rate distribution is the feature set that the subsequent unsupervised sound clustering layer learns from. Reflecting this, harmonic neuron output signal routing is based on the presynaptic logic described in Section \ref{sec:presynaptic_routing}, with two notable differences.

First, signal routing is proportional to pressure raised to the fourth power. This amplifies the signal-pressure feedback loop, increasing signal concentration towards predictive neurons. Additionally, output signals are split into distinct exploratory and activating components to protect non-predictive neuron synapse strengths, as introduced in Section \ref{sec:signal_types}. Both an exploratory and activating copy of each output signal are created, and the neuron independently distributes each version of the signal correspondingly to similarly separated exploratory and activating pressure values in each synapse (Eq. \ref{eq:harmonic_routing_exploratory}, \ref{eq:harmonic_routing_activating}).

\begin{equation}
    SIe_{i} = SIe * \frac{PRe_{i}^{4}}{\sum{}_{j}PRe_{j}^{4}} 
    \label{eq:harmonic_routing_exploratory}
\end{equation}

\begin{equation}
    SIa_{i} = SIa * \frac{PRa_{i}^{4}}{\sum{}_{j}PRa_{j}^{4}} 
    \label{eq:harmonic_routing_activating}
\end{equation}

\subsection{Sound Neurons} \label{sec:sound_neurons}
The fourth layer contains sound neurons that perform clustering by learning harmonic profiles. When the network perceives a stimulus such as a clarinet for the first time, a previously uncommitted sound neuron will learn the clarinet’s profile, return to dormancy when the clarinet stops, and reactivate later when another clarinet is perceived. 

Sound neurons build on the core postsynaptic SLP methods described in Sections \ref{sec:postsynaptic_prediction} and \ref{sec:emergent_learning} by additionally incorporating maturity, signal splitting, and lateral coordination as described in Sections \ref{sec:maturity}, \ref{sec:signal_types}, and \label{sec:coordinator}. We can build a full computational picture of sound neuron behavior by starting from baseline pressure and strength logic and adding each supplementary mechanism incrementally.

\subsubsection{Incorporating Maturity} \label{sec:incorporating_maturity}
Baseline pressure logic removes pressure equal to a quantity of received signal from the receiving synapse, and redistributes this pressure across all synapses proportionally to strength (Eq. \ref{eq:pressure_removal_base}, \ref{eq:pressure_allocation_base}).

\begin{equation}
    PR_{i} \leftarrow{} PR_{i} - SI_{i}\\
    \label{eq:pressure_removal_base}
\end{equation}

\begin{equation}
    PR_{j} \leftarrow{} PR_{j} + SI_{i} * \frac{ST_j}{\sum_{k}ST_{k}}
    \label{eq:pressure_allocation_base}
\end{equation}

This method is effective for stimulus patterns that have already been learned. However, when an undeveloped neuron first imprints on a stimulus, disparities between stimulus and strength profiles can cause strongly committed yet incorrect pressure predictions that drive away stimulus signals. To counteract this, the amount of redistributed pressure for each received signal is scaled by a neuron’s maturity (Eq. \ref{eq:maturity}, \ref{eq:pressure_removal_maturity}, \ref{eq:pressure_allocation_maturity}). This results in weaker predictions and greater plasticity for developing neurons, and stronger and more stable predictions for mature neurons.

\begin{equation}
    M = \frac{\sum{}_{i}ST_{i}}{1000}
    \label{eq:maturity}
\end{equation}

\begin{equation}
    PR_{i} \leftarrow{} PR_{i} - (SI_{i} \cdot{} M)\\
    \label{eq:pressure_removal_maturity}
\end{equation}

\begin{equation}
    PR_{j} \leftarrow{} PR_{j} + (SI_{i} \cdot{} M) * \frac{ST_j}{\sum_{k}ST_{k}}
    \label{eq:pressure_allocation_maturity}
\end{equation}

Maturity is also used to gradually slow learning and stabilize representations as neurons develop by scaling the baseline strength increment (Eq. \ref{eq:strength_update_base}) by inverse maturity (Eq. \ref{eq:strength_update_maturity}).

\begin{equation}
    ST_{i} \leftarrow{} ST_{i} + SI_{i}
    \label{eq:strength_update_base}
\end{equation}

\begin{equation}
    ST_{i} \leftarrow{} ST_{i} + (SI_{i} \cdot{} (1 - M))
    \label{eq:strength_update_maturity}
\end{equation}

\subsubsection{Incorporating Split Signal Types} \label{sec:incorporating_split_signal_types}
To fully prevent synaptic strength changes for non-predictive neurons, the sound layer incorporates exploratory and activating signal splitting as introduced in Section \ref{sec:signal_types} by creating separate pressure and signal processing pathways for each signal type.

Pressure redistribution for the exploratory pathway functions as described above, but received exploratory signals do not modify synaptic strengths. Instead, cumulative average exploratory input signal quantity \textit{\textbf{isqe}} is tracked using the isq equation provided in section \ref{sec:pitch_neurons}. Synapse activating pressures are set to 0 unless isqe is above a threshold, in which case they are set equal to exploratory pressures. This results in harmonic neurons distributing exploratory signals across all sound neurons, but routing strength-modifying activating signals only to neurons with high isqe. 

Thus, sound neuron exploratory logic encompasses pressure equations \ref{eq:pressure_removal_exploratory} and \ref{eq:pressure_allocation_exploratory}, and isqe equation \ref{eq:isqe}, while activating logic encompasses pressure equation \ref{eq:pressure_activating} and strength update equation \ref{eq:strength_update_activating}.

\begin{equation}
    PRe_{i} \leftarrow{} PRe_{i} - (SIe_{i} \cdot{} M)\\
    \label{eq:pressure_removal_exploratory}
\end{equation}

\begin{equation}
    PRe_{j} \leftarrow{} PRe_{j} + (SIe_{i} \cdot{} M) * \frac{ST_j}{\sum_{k}ST_{k}}
    \label{eq:pressure_allocation_exploratory}
\end{equation}

\begin{equation}
    isqe_{t+1} = (decay * isqe_{t}) + ((1 - decay) * SIe_{t})
    \label{eq:isqe}
\end{equation}

\begin{equation}
    PRa_{j} = 
    \begin{cases} 
      PRe_{j} & \text{if } isqe > threshold \\
      0 & \text{otherwise}
    \end{cases}
    \label{eq:pressure_activating}
\end{equation}

\begin{equation}
    ST_{i} \leftarrow{} ST_{i} + (SIa_{i} \cdot{} (1 - M))
    \label{eq:strength_update_activating}
\end{equation}

\subsubsection{Incorporating Lateral Coordination} \label{sec:incorporating_lateral_coordination}
The final sound layer feature is a coordinator node (as introduced in Section \ref{sec:coordinator}) that limits activating signal reception to a single neuron at a time, so that neurons can be treated as mutually exclusive classes and activating signal routing as a clustering outcome.

In RISP, the coordinator is implemented as a standalone layer with a single neuron that has only output synapses. Each synapse connects to a sound neuron, which sets synaptic exploratory pressure to its isqe value (Eq. \ref{eq:coordinator_pressure_exploratory}). The coordinator sets activating pressure for the highest isqe synapse to 1, and to 0 for all other synapses (Eq. \ref{eq:coordinator_pressure_activating}). Finally, sound neuron activating pressure gating is extended to require 1-valued pressure from the coordinator synapse (Eq. \ref{eq:pressure_activating_coordinated}).

\begin{equation}
    PRe_{c} \leftarrow{} isqe
    \label{eq:coordinator_pressure_exploratory}
\end{equation}

\begin{equation}
    PRa_{i} = 
    \begin{cases} 
      1 & \text{if } PRe_{i} = \max_{s \in Synapses} PRe(s) \\
      0 & \text{otherwise}
    \end{cases}
    \label{eq:coordinator_pressure_activating}
\end{equation}

\begin{equation}
    PRa_{j} = 
    \begin{cases} 
      PRe_{j} & \text{if } isqe > thresh \text{ and } PRa_{c} = 1 \\
      0 & \text{otherwise}
    \end{cases}
    \label{eq:pressure_activating_coordinated}
\end{equation}

\section{Results}
\begin{figure}
    \centering
    \includegraphics[width=1\linewidth]{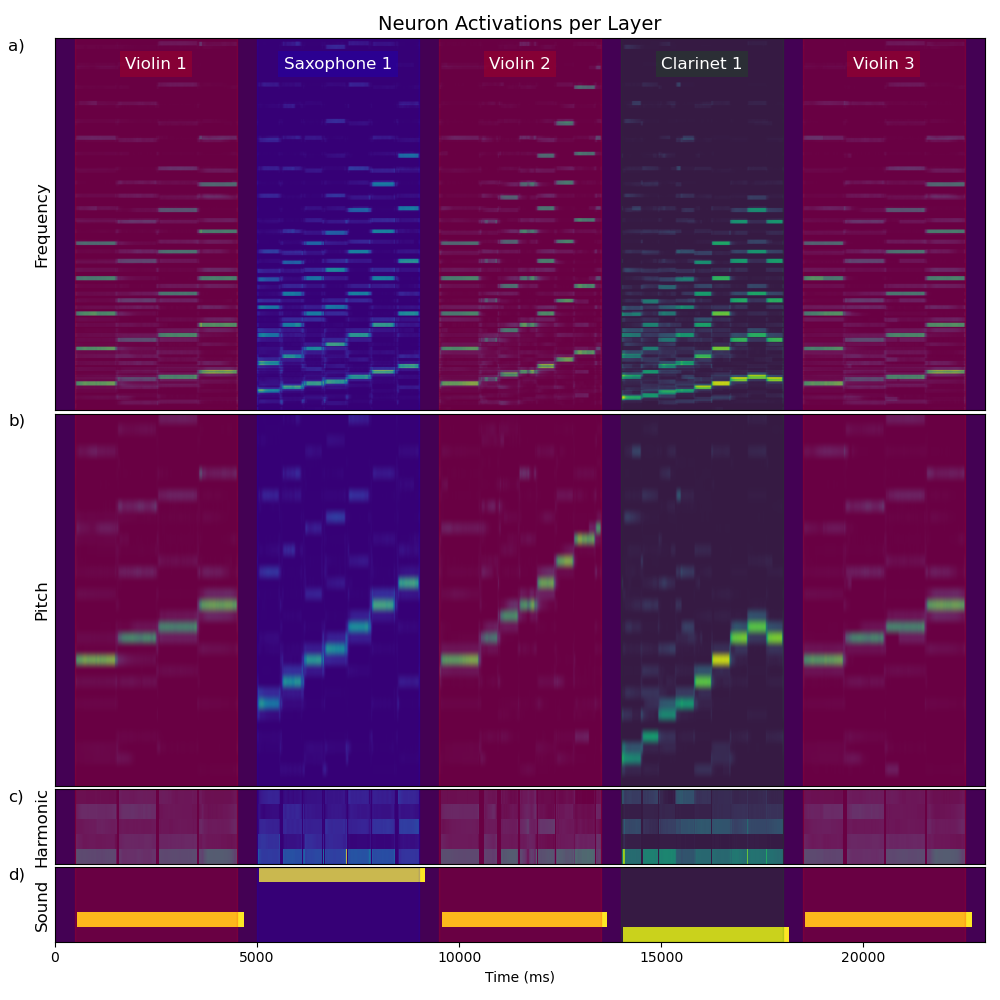}
    \caption{Neural activations across network layers during a 23-second audio stream containing 5 instrument scales. a) Frequency layer activity reflects input audio spectral amplitudes. b) Pitch neurons consolidate activity across harmonic series stimulus frequencies. c) Harmonic neuron activity forms a pitch-invariant harmonic profile that is broadly consistent for each instrument. d) The sound layer produces a single active neuron at a time, corresponding to a cluster assignment.}
    \label{fig:activations}
\end{figure}

The RISP SLP implementation was evaluated for its ability to support a representative set of heterogeneous SLP mechanisms in a task-driven network, as well as for its computational efficiency relative to the original SLP codebase. For the network’s task, we designed an unsupervised auditory learning experiment in which the network processes audio streams containing sequences of scales played by unknown mixtures of instruments, and produces a cluster assignment for each time step (Fig Fig \ref{fig:activations}a, \ref{fig:activations}d).

Audio samples were derived from 4-second clips of 24 scales from \textit{Good-sounds} dataset~\cite{bandiera2016good}, played by tenor saxophone, clarinet, flute, and violin. These clips were used to construct 100 23-second tracks, each containing 5 randomly ordered scales separated by 0.5 seconds of silence. Each track was transformed using a short-time Fourier transform (50\,ms window, 8\,ms stride), producing frequency amplitudes from 10Hz to 24kHz in 10Hz increments which were streamed into the network as frequency layer input.

Clustering is performed by the sound layer, where each neuron is treated as a cluster, and each time point as a data point with an unseen ground-truth label corresponding to the instrument in its originating audio segment. A time point is assigned to whichever (if any) neuron has its activating pressure and signal pathway open.

Two complementary metrics were used to measure accuracy at different levels of temporal resolution. For coarse-grained analysis, we treat each of the 5 instrument clips as a single data point assigned to the most frequently occurring cluster across its time steps, and measure Normalized Mutual Information (NMI) which computes a [0, 1] overlap score with the hidden clip labels. For fine-grained analysis, we calculate [0, 1] Confidence as the fraction of time steps within each clip assigned to its primary cluster, averaged across clips.

As a baseline, we compared against the DBSCAN clustering algorithm, which similarly does not require prior knowledge of the number of clusters. Harmonic neuron activations were logged for each time point to construct feature vectors equivalent to those available to the sound layer, and DBSCAN was applied offline. Hyperparameters for both methods were tuned to maximize a combined objective of NMI and Confidence.

\begin{figure}
    \centering
    \includegraphics[width=1\linewidth]{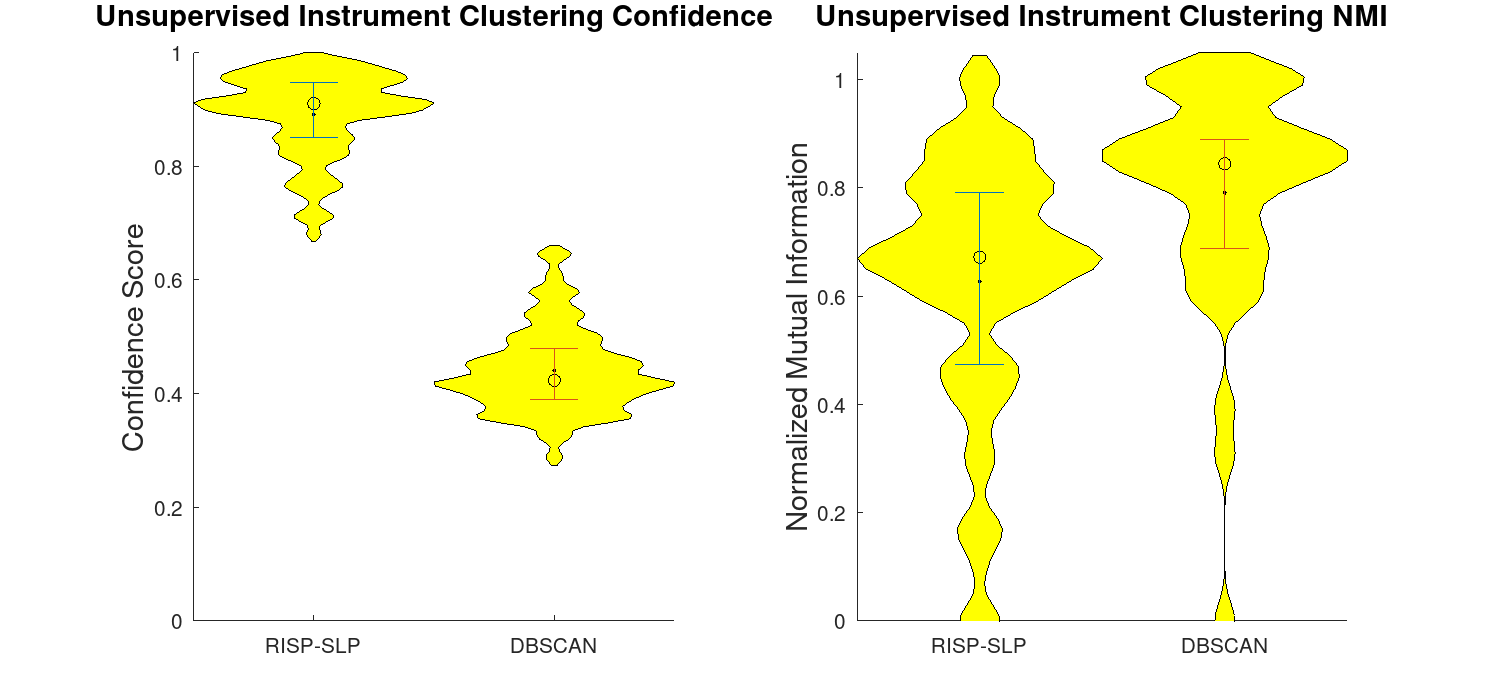}
    \caption{Comparing Normalized Mutual Information for both algorithms}
    \label{fig:nmi}
\end{figure}

Results are summarized in Figure ~\ref{fig:nmi}. DBSCAN achieved a higher NMI of 0.79 compared to 0.63 for SLP, but a substantially lower Confidence of 0.45 compared to 0.89 for SLP, producing multiplicative combined scores of 0.36 for DBSCAN and 0.56 for SLP. This suggests that DBSCAN has higher coarse-grained accuracy in correctly assigning clip-level clusters more often, while SLP has higher fine-grained accuracy in correctly assigning a higher total percentage of individual time points.

Finally, we compared processing time between the RISP SLP implementation and the original SLP codebase, using an identical subset of input audio tracks. RISP achieved a processing time of 17.2\,s per 23-second track, compared to 884.6\,s for the original implementation, corresponding to a 51$\times$ speedup. Notably, RISP processing speed was also 1.3$\times$ real time, crossing a critical threshold required for SLP to perform live processing tasks.

\section{Conclusion}
This work demonstrates that the SLP framework can be effectively implemented within a neuromorphic simulation environment using the RISP neuroprocessor. The resulting system preserves the core SLP mechanisms of bidirectional, localized synaptic feedback and neuron model heterogeneity, and exhibits emergent learning that achieves performance comparable to the DBSCAN algorithm on a musical instrument clustering task. The RISP-based implementation also achieves substantial efficiency gains, increasing processing speed by over 50$\times$ and enabling real-time processing. Together, these results establish the feasibility of translating SLP mechanisms into neuromorphic systems and highlight RISP as a suitable platform for heterogeneous SNN architectures.

More broadly, these findings suggest that population-level learning dynamics can be explicitly engineered within neuromorphic systems maintaining strictly localized signaling. They also demonstrate the flexibility of neuromorphic architectures to support heterogeneous computation and communication patterns tailored to specific tasks.

However, important limitations remain. The instrument clustering task is a relatively constrained problem compared to real-world learning scenarios, which typically involve more complex structure, increased noise, and often some form of feedback or supervision. Applying SLP to more challenging tasks is therefore necessary to assess its broader applicability. In addition, RISP software support for SLP mechanisms is not yet translatable into hardware, which limits the neuromorphic efficiency benefits available for SLP implementations.

Future work will focus on scaling SLP networks to more complex applications such as blind sound source separation (BSSs), which requires learning, tracking, and separating multiple concurrent sounds in dynamic auditory environments. This will involve expanding pitch neuron models, adding source-tracking layers, and incorporating user feedback pathways. 

A longer-term challenge is translating SLP mechanisms to hardware. Existing neuromorphic platforms are typically optimized for LIF-based neuron and synapse models and often do not natively support key SLP features such as graded spikes, bidirectional synapses, and multiple signal types. Exploring hardware designs that accommodate these mechanisms will be essential for fully realizing SLP in neuromorphic systems. In the near term, however, the processing speed gains observed in the RISP SLP implementation suggest that software-based neuromorphic simulation is a viable path for deploying SLP for real-time tasks for the near future.

Overall, this work represents a step toward expanding the design space of neuromorphic systems to include heterogeneity and engineered emergence as flexible tools for constructing effective task-adaptive network designs. Further investigation will determine whether these approaches, combined with the efficiency advantages of neuromorphic computation, can provide a viable alternative to data- and compute-intensive deep learning methods for complex real-world tasks.

\bibliographystyle{IEEEtran}
\bibliography{bib}

\end{document}